\documentclass[journal]{IEEEtran}
\usepackage{cite}
\usepackage{placeins}
\usepackage{CJKutf8}
\usepackage{float}
\usepackage{amsmath,amssymb,amsfonts}
\usepackage{algorithmic}
\usepackage{graphicx}
\usepackage{textcomp}
\usepackage{multirow}
\usepackage{caption}
\usepackage{placeins}
\usepackage{url}
\usepackage{hyperref}
\usepackage{tabularx}
\usepackage{multirow}
\usepackage{array}
\usepackage{booktabs}
\usepackage{hhline}
\usepackage{makecell}
\usepackage{CJKutf8}
\begin{document}
\captionsetup{justification=centering}
\title{Deep Learning-Based Feature Fusion for Emotion Analysis and Suicide Risk Differentiation in Chinese Psychological Support Hotlines}
\author{Han Wang, Jianqiang Li, Qing Zhao, Zhonglong Chen, Changwei Song, Jing Tang, Yuning Huang, Wei Zhai, Yongsheng Tong\textsuperscript{*}, Guanghui Fu\textsuperscript{*}
\thanks{Han Wang, Jianqiang Li, Qing Zhao, Zhonglong Chen, Changwei Song, Jing Tang, Wei Zhai are with School of Software Engineering, Beijing University of Technology, Beijing, China.}
\thanks{Yongsheng Tong is with Peking University Huilongguan Clinical Medical School, Beijing, China; WHO Collaborating Center for Research and Training in Suicide Prevention, Beijing, China.}
\thanks{Guanghui Fu is with Sorbonne Universit\'{e}, Institut du Cerveau – Paris Brain Institute - ICM, CNRS, Inria, Inserm, AP-HP, H\^{o}pital de la Piti\'{e}-Salp\^{e}tri\`{e}re, Paris, France (e-mail: guanghui.fu@inria.fr).}
\thanks{Corresponding author: Guanghui Fu (\url{guanghui.fu@inria.fr})}
\thanks{This work was supported by The National Natural Science Foundation of China under Grant No.82071546.}
}
\maketitle
\begin{abstract}
Mental health is a significant global public health issue, and psychological support hotlines play a crucial role in providing mental health assistance and identifying suicide risks at an early stage. However, the emotional expressions conveyed during these calls remain underexplored in current research.  This study introduces a novel method that combines pitch acoustic features with deep learning-based features to analyze and understand emotions expressed during hotline interactions. Using data from China’s largest psychological support hotline, which includes 105 subjects, our method achieved an F1-score of 79.13\% for negative binary emotion classification. Additionally, the proposed approach was validated on an open dataset for multi-class emotion classification, where it demonstrated better performance compared to the state-of-the-art methods. To explore its clinical relevance, we applied the model to analysis the frequency of negative emotions and the rate of emotional change in the conversation, comparing 46 subjects with suicidal behavior to those without. While the suicidal group exhibited more frequent emotional changes than the non-suicidal group, the difference was not statistically significant. 
Importantly, our findings suggest that emotional fluctuation intensity and frequency could serve as novel features for psychological assessment scales and suicide risk prediction.
The proposed method provides valuable insights into emotional dynamics and has the potential to advance early intervention and improve suicide prevention strategies through integration with clinical tools and assessments.
The source code is publicly available at: \url{https://github.com/Sco-field/Speechemotionrecognition/tree/main}.
\end{abstract}
\IEEEpeerreviewmaketitle

\section{Introduction}
\IEEEPARstart {M}{ental}{M}{ental} health is a critical component of overall well-being, influencing an individual's ability to cope with stress, relate to others, and make decisions. However, when mental health disorders go unaddressed, they can lead to severe consequences, including suicide. is one of the major social and public health issues globally, with approximately 804,000 people losing their lives to suicide every year 
~\cite{o2018integrated}.China is responsible for 15\% of the total global suicide deaths~\cite{cheng2020suicide}. Each death is a tragedy for the family members, friends, and colleagues of the deceased, but suicide is preventable. Early identification of suicide risk and timely intervention are crucial for reducing suicide rates~\cite{fonseka2019utility}. As a result, many countries, both developed and developing, have established nationwide networks of psychological support hotlines~\cite{zalsman2021suicide}. These hotlines have been proven to be important and highly effective tools for suicide prevention\cite{gould2018follow, gould2016helping}. Language is not only a basic means of communicating words but also an important way to convey a wide range of human emotions~\cite{lotfidereshgi2017biologically}. Callers to psychological support hotlines often express a range of negative emotions through their speech~\cite{shaw2019evaluation}. Identifying these negative emotions is critical for timely intervention~\cite{nfissi2024unlocking}, as it helps professionals quickly assess the caller's psychological state and take appropriate actions. By analyzing the content of the callers' speech, hotline staff can detect emotions such as anger, sadness, and fear~\cite{chen2024fine}
, which allows them to provide more effective emotional support and practical assistance.

With the advancement of technology, Speech Emotion Recognition (SER) has emerged as a new approach focusing on the automatic detection and recognition of emotions in speech signals~\cite{hashem2023speech}. In recent years, this field has gained significant attention, becoming a promising method for monitoring mental health and well-being, especially for individuals at potential risk of suicide~\cite{chen2024fine}
. Thus, SER-related technologies can be utilized to better identify the emotions of patients in psychological support hotline, thereby providing more effective emotional support and intervention. This technology can help hotline staff understand callers' emotional states in real-time, enabling them to respond promptly and adjust intervention strategies based on different emotional states, thereby increasing the success rate of intervention and reducing suicide risk. By analyzing emotional cues in speech, staff can provide the necessary assistance and resources. Chen et al ~\cite{chen2024fine} 
explored the effectiveness of various large-scale pre-trained models in emotion recognition for psychological support hotline, while Song et al.~\cite{song2024exploratory} used Whisper for feature extraction and incorporated positional embedding to assess the risks in psychological support hotline. Gideon et al.~\cite{gideon2019emotion} collected a dataset focused on studying the relationship between suicidal ideation and emotional fluctuations, consisting of voice recordings and self-reports from individuals who had experienced suicidal thoughts or attempts, and used eGeMAPS features~\cite{eyben2015geneva} to identify the emotional states of patients. Cui et al.~\cite{cui2024spontaneous}combined Whisper~\cite{radford2023robust} with large language models (LLMs), employing different fine-tuning strategies for both speech and text models, and tested various fusion methods to integrate speech and text modalities. These studies collectively demonstrate the significant value of SER technology in suicide risk detection. However, these studies have not fully utilized various audio features. Effectively exploring and applying multiple audio features could further improve the effectiveness of suicide risk detection. By integrating different acoustic features (such as spectrogram features and MFCC features), more comprehensive emotional information can be obtained, which helps more accurately identify potential suicide risks. This multi-feature fusion approach contributes to significantly enhancing the overall performance of the model, enabling more effective intervention and support.

Studies have shown that effectively combining features extracted from the Transformer architecture with raw audio features can significantly improve model performance. This feature fusion strategy takes full advantage of the Transformer's strengths in capturing contextual information while retaining subtle characteristics of the original audio signal, thereby achieving more accurate emotion recognition and risk detection. Zou et al.~\cite{zou2022speech} utilized spectrogram features, MFCC features, and features extracted from Wav2Vec2 for concatenated fusion. Liu et al.~\cite{liu2023dual} fused features extracted from Opensmile with those extracted using Wav2Vec2. However, these studies did not fully consider the integration of tonal features, especially in tonal languages, where tone, as an important feature, can further enhance the accuracy of emotion recognition.tIn this study, we combined deep learning features from the pretrained Wav2Vec2 model with pitch and MFCC features to create a merged representation. We then utilized an attention mechanism for feature fusion to classify emotions in speech recordings from a psychological support hotline. The data were sourced from China's largest psychological support hotline from Beijing Huilongguan Hospital, with a total of 20,630 segments from 105 subjects. To this end, we built a binary emotion recognition model aimed at identifying the emotions of callers. Experimental results showed that the introduction of merged features improved the accuracy of emotion recognition. Our model demonstrated strong performance in the context of a psychological support hotline, achieving an F1-score of 79.13\% and outperforming seven other comparison models in this metric. 
We also evaluated our method on a Vietnamese dataset for emotion recognition~\cite{thanh2024robust}, where it outperformed state-of-the-art methods. Additionally, we applied this tool to analyze the emotional trends of 46 subjects, comparing suicidal and non-suicidal groups. The results showed that individuals in the suicidal group exhibited a higher frequency of negative emotions and pronounced emotional instability. Specifically, they experienced more frequent emotional fluctuations and prolonged negative emotional states—key indicators of the emotional volatility commonly associated with suicidal ideation. While these differences were not statistically significant, the findings reveal critical patterns in emotional dynamics that hold substantial clinical relevance.
The observed emotional fluctuations and prolonged negative states in the suicidal group suggest that these patterns could serve as novel biomarkers for suicide risk assessment. Unlike traditional static metrics, such as one-time psychological scale scores, emotional trend analysis provides dynamic insights into an individual’s mental state over time. This approach captures real-time emotional instability and distress, often overlooked in conventional methods, and could complement existing psychological assessment tools. Moreover, integrating emotional fluctuation metrics into psychological scales or predictive models has the potential to improve the precision and sensitivity of suicide risk assessments, enabling more targeted and timely interventions.
These findings highlight the broader clinical value of emotion analysis in psychological support settings, offering a promising direction for future research. By leveraging these insights, psychological support hotlines and mental health professionals could better understand an individual’s emotional trajectory, ultimately enhancing early detection and prevention of suicide.

\section{Related work} \label{sec:related} 

Speech Emotion Recognition (SER) aims to identify the emotional state in human speech signals~\cite{shen2024emotion}, and it is now applied in various fields, such as disease detection~\cite{khan2023deep}, system safety~\cite{saste2017emotion}, emergency protocols~\cite{deschamps2022investigating}, and suicide risk detection. Its process is generally divided into two main steps: feature extraction and classification~\cite{koolagudi2012emotion}.

The selection and extraction of audio features are crucial in tasks such as negative emotion recognition. Different audio features capture various attributes of the speech signal, providing the foundation for machine learning models to classify and predict based on the speech signal.

MFCC (Mel-frequency cepstral coefficients) is a low-level speech feature widely used in Speech Emotion Recognition (SER) and contains rich emotion-related information~\cite{liu2022speech}. It is based on the human ear's perception of frequency and processes speech signals through a nonlinear transformation. MFCC has been successfully applied to emotion recognition and speaker identification. Typically, MFCC is used in combination with its first- and second-order derivatives to capture the temporal variations in speech signals, thereby improving the accuracy of emotion recognition.In light of this, Abubakar et al.~\cite{abubakar2024stutternet} extracted MFCC spectrogram features from stuttering patients using BiLSTM, and combined these temporal features to detect stuttering events in the patients' speech. Their approach effectively utilized the dynamic characteristics of MFCC, capturing the temporal variations in speech signals, and achieved significant results in the task of stutter detection. This research demonstrates that the combination of MFCC features and BiLSTM is highly effective.

At the same time, with the advancement of deep learning technology, models based on the Transformer architecture~\cite{vaswani2017attention}, such as Wav2Vec2~\cite{baevski2020wav2vec}, HuBERT~\cite{hsu2021hubert}, and Whisper~\cite{radford2023robust} , have gradually emerged as new methods for speech feature extraction. Wav2Vec2~\cite{baevski2020wav2vec} is a self-supervised learning model based on the Transformer architecture, specifically designed for speech processing tasks. Through self-supervised learning, it extracts high-level speech representations from a large amount of unlabeled speech data. The model first uses convolutional neural networks to extract low-level features and then undergoes pre-training through masked prediction, utilizing contextual information to predict the masked portions, thereby learning the contextual dependencies of speech signals. The Transformer network models long-range dependencies in speech and extracts complex patterns. The pre-trained Wav2Vec2 can be fine-tuned for tasks such as speech recognition, emotion recognition, and speaker identification, demonstrating excellent generalization ability.

For tonal languages, the role of tonal features in speech signals is particularly critical. Tones not only affect word meanings but also convey rich emotional information. Therefore, relying solely on traditional MFCC or high-level features from self-supervised learning may struggle to accurately capture the subtle nuances of tone in emotional expression. Especially in tonal languages like Chinese and Vietnamese, syllables can have entirely different meanings based on their tones. Although some pioneering studies have proposed specialized frameworks for speech processing tasks in tonal languages~\cite{kaur2021automatic,huu2023mispronunciation},  research in the field of emotion recognition specifically for tonal languages remains relatively limited. Thanh et al.~\cite{thanh2024robust} achieved excellent results by utilizing tonal features and features extracted from Wav2Vec2 on Chinese, Thai, and a self-created Vietnamese dataset, which also inspired our work.

Existing research indicates that MFCC features and Transformer-based models, such as Wav2Vec2, play important roles in speech emotion recognition~\cite{saadati2024multi}. MFCC effectively captures the temporal variations in speech signals, while self-supervised learning models like Wav2Vec2 excel in high-level feature extraction and long-range dependency modeling. 
However, research on tonal languages is relatively limited, especially in capturing the subtle influences of tone on emotional expression, indicating that there is still room for improvement.

\section{Methods}

\begin{figure*}[ht]
    \centering
    \includegraphics[width=\textwidth]{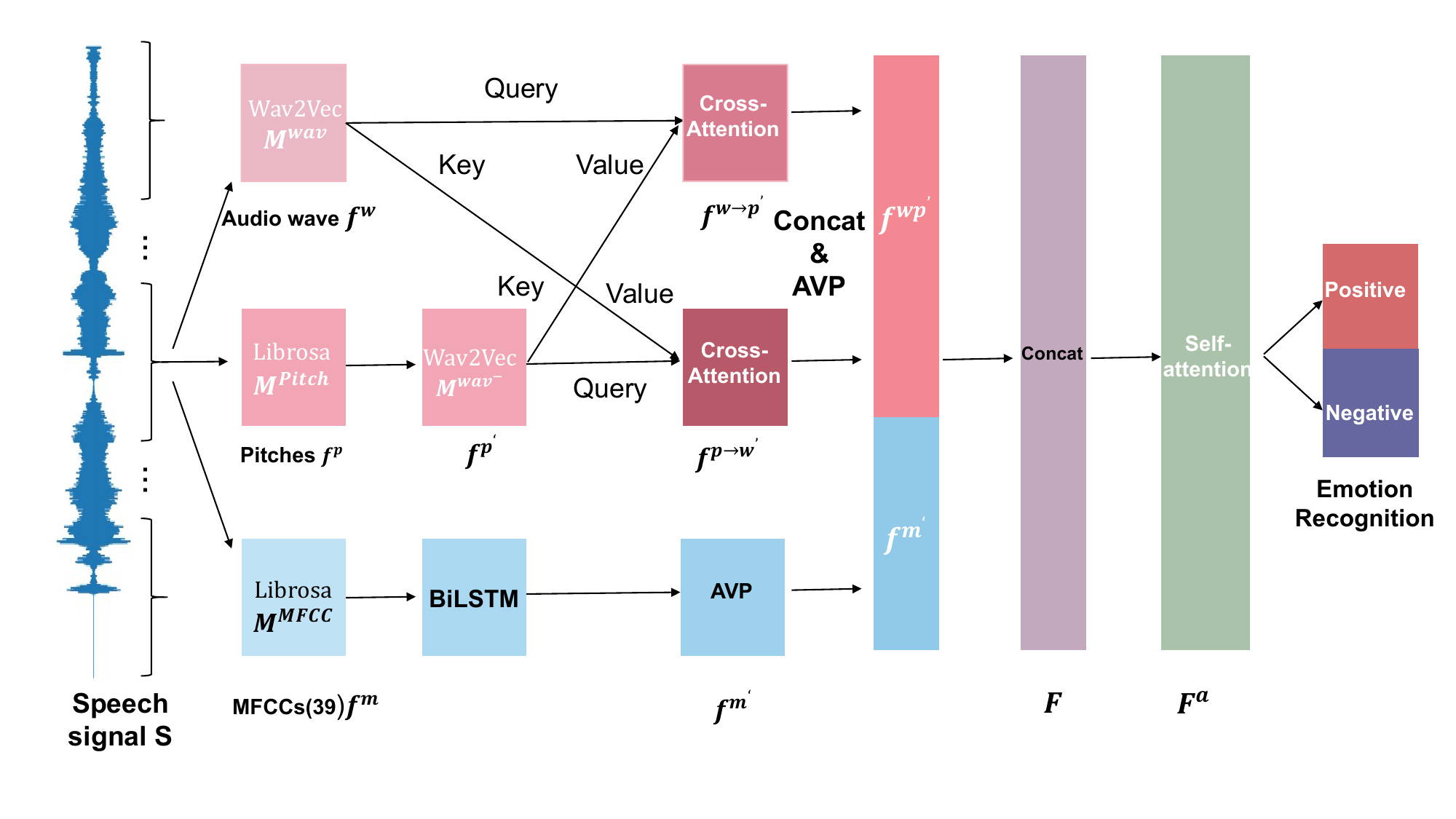} 
    \caption{The model architecture of the proposed method.}
    \label{fig:model_architecture}
\end{figure*}

In this section, we describe a speech emotion recognition (SER) system based on a joint attention mechanism, which achieves emotion classification by integrating multiple speech features. Figure~\ref{fig:model_architecture} illustrates the overall framework of our proposed method. As shown in the figure, we first extract three hierarchical speech features (tonal features, MFCC features, and Wav2Vec2 features), which are processed through their respective feature encoding networks. Finally, feature fusion is performed through the attention mechanism for classification.

\subsection{Feature extraction}
We extract features from the audio using three methods: deep learning features obtained from the pre-trained Wav2Vec backbone~\cite{baevski2020wav2vec}, pitch features that account for the tonal characteristics of the language, and MFCC (Mel-Frequency Cepstral Coefficient) features. For a given speech signal $S$, consisting of $n$ audio segments $s_1, s_2, ..., s_n$, each segment was processed using three feature extractor: pre-trained Wav2Vec backbone $M^\mathrm{wav}$which extracts the deep learning feature $f^w$; the pitch feature extractor $M^\mathrm{pitch}$, which extracts the tonal feature $f^p$; and the MFCC extractor $M^\mathrm{MFCC}$, which generates the MFCC feature $f^m$, as described in Equation~\ref{eq:feature_extract_s1}.
\begin{equation} \label{eq:feature_extract_s1}
\begin{aligned}
    f^{w} &= M^{\mathrm{wav}}(s_n) \\
    f^{p} &= M^\mathrm{pitch}(s_n) \\
    f^{m} &= M^\mathrm{MFCC}(s_n)
\end{aligned}
\end{equation}
Wav2Vec is effective in extracting features from raw audio signals. However, due to the tonal nature of the Chinese language and the specific characteristics of a psychological support hotline, tonal changes are crucial as they often reflect emotional shifts. To address this, we combine the pitch feature $f^p$ with the original Wav2Vec feature $f^w$ leveraging tonal information while retaining the raw audio features. To achieve this, we apply the encoder-decoder architecture of the Wav2Vec model $M^{\mathrm{wav}^-}$to $f^p$, ensuring the output matches the dimensionality of $f_w$. Note that the $M^{\mathrm{wav}^-}$used here does not utilize pre-trained weights. The processed pitch feature representation is denoted as $f^{p\prime}$. 
Compared to deep learning-based features, MFCC features are directly extracted from the audio spectrum, capturing the speech characteristics of the audio signal. Specifically, 13 MFCC coefficients are calculated, along with the first- and second-order derivatives, resulting in a 39-dimensional feature representation. These features are then encoded using a bidirectional LSTM, $M^\mathrm{BiLSTM}$, followed by dimensionality reduction through average pooling $AVP$ to obtain the reduced feature representation, $f^{m\prime}$. These process can be represent as Equation~\ref{eq:feature_extraction_s2}:
\begin{equation} \label{eq:feature_extraction_s2}
    \begin{aligned}
        f^{p\prime} &= M^{\mathrm{wav}^-} (f^{p}) \\
        f^{m\prime} &= \mathrm{AVP}(M^{\mathrm{BiLSTM}}(f^m))
    \end{aligned}
\end{equation}

\subsection{Feature fusion}
As described earlier, our approach combines the deep learning feature $f^w$ with the pitch feature $f^{p'}$ to obtain pitch-enhanced features. This is achieved through the use of cross-attention mechanisms, as shown in Equation~\ref{eq:feature_fusion}:

\begin{equation} \label{eq:feature_fusion}
\begin{aligned}
    f^{w \to p'} &= \text{CrossAttention}(f^w, f^{p'}) \\
    f^{p' \to w} &= \text{CrossAttention}(f^{p'}, f^w)
\end{aligned}
\end{equation}

In this formulation, $f^{w \to p'}$ represents the feature obtained by attending to the pitch feature $f^{p'}$ using the deep learning feature $f^w$, while $f^{p' \to w}$ represents the feature obtained by attending to the deep learning feature $f^w$ using the pitch feature $f^{p'}$.
Next, the two features are merged using concatenation followed by an average pooling operation, $AVP$, to obtain the merged feature $f^{wp\prime}$, as shown in Equation~\ref{eq:feature_fusion_s2}. 
\begin{equation} \label{eq:feature_fusion_s2}
f^{wp\prime} = \text{AVP}(\text{Concat}(f^{w \to p'}, f^{p' \to w}))
\end{equation}
The final feature $F$ is obtained by combining $f^{wp\prime}$ and $f^{m\prime}$ through concatenation, as shown in Equatio~\ref{eq:feature_fusion_s3}. 
\begin{equation} \label{eq:feature_fusion_s3}
F = \text{Concat}(f^{wp\prime}, f^{m\prime})
\end{equation}

\subsection{Classifier and loss function}
The feature $F$ describe before is the combination of three type of feature, we use self attention operation to let them build the connection between these features.
\begin{equation}
    F^{a} = \text{SelfAttention}(F) \\
\end{equation}
Then the resulting features $F^a$ are passed through a classifier designed for the downstream classification task. The classifier begins by applying a Dropout layer to the fused features, which helps reduce the risk of overfitting and improves the model's generalization ability. Following the Dropout layer, the input features are transformed through a fully connected layer, generating a new representation of the features in the output space, enabling the classifier to learn from the fused features. The Thanh activation function is then applied.
To further avoid overfitting and ensure good generalization to unseen data, a second Dropout layer is used. The classifier outputs the class probabilities using a Softmax function for multi-class classification or a Sigmoid function for binary classification, providing the final prediction for each sample.
For the loss function, binary cross-entropy with logits loss (BCEWithLogitsLoss) is employed for binary classification tasks, while cross-entropy loss is used for multi-class classification tasks.

\section{Experiments} \label{sec:experiments}
\subsection{Datasets}
This method was validated using two datasets for sentiment analysis and emotion classification tasks: one from a private clinical psychological support hotline for Chinese speakers and the public Vietnamese dataset, ViSEC~\cite{thanh2024robust}. The method was primarily designed for Chinese speech emotion analysis. The Vietnamese dataset was chosen due to the similarities between Chinese and Vietnamese, as both are tonal languages that use syllables as their basic linguistic units. These characteristics align with the principles of the proposed method, which emphasizes tonal and pitch features

\paragraph{Private dataset} \label{sec:exp:data:private}
This clinical private dataset comes from the psychological support hotline at Beijing Huilongguan Hospital.
The assessment process begins with suicide risk measurement, establishing trust to gather background and psychological information, followed by using Tong et al.'s~\cite{tong2020prospective, tong2023predictive} suicide risk scales to objectively evaluate risk levels. These scales incorporate both continuous and dichotomous variables to generate an aggregate score indicating low-moderate or high suicide risk. Next, crisis intervention addresses immediate risks, with flexibility to adjust the sequence if callers exhibit high suicidal ideation. Finally, a 12-month follow-up stage ensures ongoing monitoring and support, where suicidal behavior during the follow-up period serves as the final model label for analysis. 

This dataset contains 20,630 segmented audio sentences from 105 subjects. 
These segments were annotated by three experts and categorized into various negative emotions, as detailed in Chen et al.~\cite{chen2024fine}.  
All subjects completed follow-up interviews over a 12-month period, during which suicide-related behaviors (including whether or not they attempted suicide) were confirmed. Details of the psychological support hotline are provided in Song et al.~\cite{song2024exploratory}.
This dataset has a total length of 80 hours, with the average duration of each audio segment being 7.21$\pm$7.52 seconds. The average duration of each dialogue is 45$\pm$20 minutes. In this study,  we selected 20,254 audio segments for further analysis,we focus on binary classification of negative emotions, where the dataset was restructured into two categories: negative and non-negative. The dataset includes a total of 9,774 negative emotion samples and 10,856 non-negative emotion samples.
To ensure consistency, we adopted the same data-splitting methodology as described in Chen et al.~\cite{chen2024fine}. Specifically, the 105 speech segments were divided at the patient level into training and testing sets with a 4:1 ratio, resulting in 84 subjects for training and 21 subjects for testing.. Additionally, five-fold cross-validation was performed on the training set.
The data distribution can be found in Table~\ref{tab:dataset_private}. 
\begin{table}[ht]
    \centering
    \caption{Distribution of emotion types of private dataset segments.}
    \begin{tabular}{@{} l c c c c @{}} 
        \toprule 
        Emotion type & Train & Validation & Test \\ 
        \midrule 
        None-negative       & 6891   &   1794  &   2171  \\
        negative     & 6071  &   1447  &   1880  \\

        \midrule 
        Total       & 12962  &  3241  &   4051  \\ 
        \bottomrule 
    \end{tabular}
    \label{tab:dataset_private}
\end{table}

For the emotional trends and suicide risk differentiation task, we analyzed data from an additional 46 subjects for this experiment. Among these subjects, 22 had exhibited suicidal behaviors, while the remaining 24 had no suicidal behaviors during the twelve-month follow-up period. The total duration of all dialogues was 46 hours, with an average duration of 56 ± 15 minutes per session.

\paragraph{Public dataset} This public dataset, named Visec~\cite{thanh2024robust} includes 147 Vietnamese speakers and 5,280 voice samples collected from YouTube. Each audio sample is labeled with four types of information: emotion, gender, Vietnamese dialect, and speaker identity. The total duration of the dataset is 3.18 hours, and all samples are uniformly resampled to a frequency of 16,000 Hz. The content covers various scenarios, such as movies, podcasts, and game shows, with speech categorized into four emotions: happy, neutral, sad, and angry. To ensure consistency, we employed the same dataset partitioning method. Specific partitioning details can be found in Table~\ref{tab:dataset_public}. 
In terms of evaluation metrics, we used weighted average (WA) and unweighted average (UA) as assessment criteria.
\begin{table}[ht]
    \centering
    \caption{Distribution of emotion types of public dataset segments.}
    \begin{tabular}{@{} l c c c c @{}} 
        \toprule 
        Emotion type & Train & Validation & Test \\ 
        \midrule 
        Happy       & 982   &   123  &   123  \\
        Neutral     & 1206  &   151  &   150  \\
        Sad     & 863   &   108  &   108  \\
        Angry    & 1173   &   146  &   147  \\
        
        \midrule 
        Total       & 3030  &   528  &   528  \\ 
        \bottomrule 
    \end{tabular}
    \label{tab:dataset_public}
\end{table}

\subsection{Data preprocessing}
For both datasets, we applied the same preprocessing steps to standardize the audio clips. Specifically, we supplemented audio clips shorter than 10 seconds for both datasets to ensure they reached a length of 10 seconds. For audio clips longer than 10 seconds, we truncated them to limit their length to 10 seconds. All audio samples were uniformly resampled to a frequency of 16,000 Hz. This choice of processing is mainly due to considerations of GPU memory limitations, and it also helps to better align the audio segments, thereby improving the efficiency and effectiveness of model training. By standardizing the length of the audio, we can reduce input variability and ensure that the model can learn features more stably during training, thus enhancing the accuracy of emotion classification.

\subsection{Comparison models}
For the experiment on our private dataset, we compared our method with several pre-trained models and prior work to evaluate performance in the context of our task. Specifically, we tested the following models: Wav2Vec2~\cite{baevski2020wav2vec}, HuBERT~\cite{hsu2021hubert}, and Whisper~\cite{radford2023robust}. For Whisper, we further explored several versions to evaluate their performance across different configurations, including Whisper-small, Whisper-small-Chinese-base, Whisper-medium and Whisper-large-v3.The Whisper-small-Chinese-base is a variant of the Whisper-small model that has been fine-tuned on the ``cmn-hans cn'' dataset from Google Fleurs~\cite{conneau2022fleursfewshotlearningevaluation}, making it specifically optimized for Chinese language tasks. In comparison, the Whisper-large-v3 model differs from the standard large version by being trained for 2.5 times longer, with additional regularization techniques applied to improve its performance. Finally, we compared the performance of these models with the method proposed by Thanh et al.~\cite{ thanh2024robust}, which utilizes both pitch features and transformer-based features for emotion recognition in speech.
For the experiment on the public dataset, we compared our method with Wav2Vec2~\cite{baevski2020wav2vec} which achieved the second-best performance in our private dataset experiment, and the method proposed by Thanh et al.~\cite{ thanh2024robust} which is recognized as the SOTA model.

\subsection{Implementation details}
Given that the dataset includes two different languages, we fine-tuned different versions of the pre-trained Wav2Vec 2.0 models. To ensure consistency, we followed the same pre-trained model versions used in the research of Chen et al.~\cite{chen2024fine} and Thanh et al.\cite{thanh2024robust}.

For the experiment on our private dataset, we used a version of Wav2Vec 2.0\footnote{\url{https://huggingface.co/jonatasgrosman/wav2vec2-large-xlsr-53-chinese-zh-cn}} that was fine-tuned on multiple Chinese speech datasets, including Common Voice 6.1~\cite{ardila2019common}, CSS10~\cite{park2019css10}, and ST-CMDS~\cite{stcmds2017}. This fine-tuning enhanced the model's relevance and effectiveness in capturing Chinese speech characteristics. 
For the experiment on the public dataset, we used a publicly available Vietnamese pre-trained Wav2Vec 2.0 model~\cite{huggingface_wav2vec2}. This model was pre-trained on 13,000 hours of Vietnamese YouTube speech data.
We use the librosa library~\cite{mcfee2015librosa} to extract MFCC and pitch features. 

The experiments were conducted using an NVIDIA 24GB RTX 4090 GPU. We set the batch size to 8 and the learning rate to 3e-5 in the training stage. The calculations were performed in 32-bit precision, with the Adam optimizer~\cite{kingma2014adam} used for optimization. And all the experiments was trained for 30 epochs. 

We conduct ablation experiments to explore the functions of different features and attention components. And, we perform comparison experiments with advanced methods on both datasets. 
We report accuracy, recall, and F1-score for the private dataset, and unweighted accuracy and weighted accuracy for the public dataset, in accordance with the references~\cite{chen2024fine, thanh2024robust}.

\subsection{Emotion trend analysis}
We applied our model to analyze emotional trends during psychological support hotline sessions, aiming to explore patterns distinguishing individuals with suicidal and non-suicidal behaviors over a 12-month follow-up period. By visualizing these emotional trends over time, we enabled a focused comparison between the two groups, providing a clearer understanding of their distinct emotional patterns. For this experiment, we selected 46 subjects from clinical settings, comprising 24 non-suicidal individuals and 22 individuals exhibiting suicidal behaviors, as detailed in Section~\ref{sec:exp:data:private}.

We developed two metrics to evaluate the emotion analysis tasks.
Using NNS (Negative Speech Segments) as a metric, we calculated the count of speech segments containing negative emotions to measure the differences between the two groups. To better capture the rate of emotional variation, we introduced a metric called the emotion change rate (ECR), as defined in Equation~\ref{eq:rate of change}.
\begin{equation} \label{eq:rate of change}
\begin{aligned}
    \text{ECR} = \frac{1}{n-1} \sum_{i=2}^{n} \left| e_i - e_{i-1} \right|
\end{aligned}
\end{equation}
This metric represents the mean absolute difference between the probability of emotion $e_i$ in segment $s_i$ and the probability of emotion $e_{i-1}$ in the pervious segment $s_{i-1}$.

Based on the workflow of the psychological support hotline~\cite{tong2023predictive,tong2020prospective}, the first 30 minutes of the conversation serve as the suicidal risk assessment stage. Therefore, we analyzed two metrics (NNS and ECR) related to emotional trend changes during both the assessment stage and the entire conversation.
These experiments allowed us to gain a more comprehensive understanding of the emotional trajectory and variations throughout the interaction. By comparing these two stages, we aimed to uncover key emotional patterns that distinguish suicidal from non-suicidal individuals and provide valuable insights into their emotional responses during the process.


\section{Results}
\subsection{Ablation experiments}
\subsubsection{Feature ablation experiments}
We conducted a feature ablation experiment on our private dataset, and the results are shown in Table~\ref{tab:exp:abalation:feature}. 
\begin{table}[!ht]
    \caption{Results of feature ablation experiments on the private dataset. $f^w$, $f^p$, and $f^m$ represent the features extracted by the pretrained Wav2Vec model, the pitch extractor, and the MFCC extractor, respectively. $F$ represents the final feature obtained by merging all these features using cross-attention and self-attention.}
    \centering
    \begin{tabular}{@{} l c c c   @{}} 
        \toprule 
        Features & Accuracy & Recall & F1-score  \\ 
        \midrule 
        Wav2Vec 2.0 ($f^w$) &\textbf{76.97\%} & 76.96\% & 76.96\% \\
        Pitch ($f^p$)     & 66.50\%  &   74.21\% &  70.36\% \\
        MFCC ($f^m$)    & 69.34\%  &   80.06\%    & 73.68\%\\
        \textbf{Merged ($F$)}     & 75.19\%   &   \textbf{87.75\%}  & \textbf{79.13\%} \\
        \bottomrule 
    \end{tabular}
    \label{tab:exp:abalation:feature}
\end{table}

The experimental result led to a clear conclusion: the best performance was achieved by combining features using the cross-attention mechanism and self-attention mechanism, resulting in the final feature set $F$.
Analyzing the individual features, we observed that Wav2Vec2 alone performed relatively well with an accuracy of 76.97\%, as it effectively extracts high-level features from raw audio signals. However, despite its strength in capturing general acoustic patterns, it did not fully capture the tonal shifts that are crucial for tasks like emotional recognition in the context of a psychological support hotline. 
On the other hand, the MFCC features ($f^m$) captured speech characteristics directly from the audio spectrum and showed an accuracy of 69.34\% with a high recall of 80.06\%. MFCCs are widely used in speech-related tasks, but in this case, their performance indicated that while they capture speech patterns well, they may not fully leverage the emotional information embedded in tonal variations, which was critical for the task at hand.
Finally, when combining these features using the cross-attention mechanism and self-attention mechanism to form $F$, we saw an improvement across all metrics, with an F1-score of 79.13\%. This demonstrates that our approach, which incorporates both the tonal information from the pitch feature and the robust audio features from Wav2Vec and MFCC, is highly effective. By leveraging the strengths of each individual feature type, our method provides a more comprehensive representation of the audio, capturing both the emotional tonal shifts and the speech characteristics necessary for accurate classification, thereby outperforming the individual feature sets. This confirms the advantage of our feature fusion approach in tasks that require attention to both speech and emotional tone.

\subsubsection{Attention ablation experiments}
We evaluated the impact of different attention mechanisms and the results are presented in Table~\ref{tab:exp:abalation:attention}.
\begin{table}[!ht]
    \centering
    \caption{Results of attention ablation experiments on the private dataset. ``Vanilla'' refers to the model without any attention operation, while ``Cross-attention'' and ``Self-attention'' refer to the models that employ only one attention operation. ``The proposed'' refers to the final model that incorporates both attention mechanisms.}
    \begin{tabular}{@{} l c c c   @{}} 
        \toprule 
        Attention & Accuracy & Recall & F1-score  \\ 
        \midrule 
        Vanilla &75.14\% & 81.07\% & 77.76\% \\
        Cross-attention     & 75.02\%  &   79.64\% &  77.36\% \\
        Self-attention   & \textbf{75.71\%}  &   84.98\%    & 78.95\%\\
        The proposed     & 75.19\%   &   \textbf{87.75\%}  & \textbf{79.13\%} \\
        \bottomrule 
    \end{tabular}
    \label{tab:exp:abalation:attention}
\end{table}

The ``Vanilla'' model, which does not employ any attention operation, served as the baseline. It achieved an F1-score of 77.76\% shows that the model, without attention, is capable of achieving decent performance, but the use of attention mechanisms is expected to enhance these results by better capturing the relationships between features.
When only cross-attention was applied, the model's F1-score slightly decreased to 77.36\%. This suggests that simply concatenating the MFCC feature $f^{m\prime} $and the deep learning feature $f^{wp\prime}$, without leveraging the relationships between them, is not an effective fusion strategy. 
In contrast, self-attention performed better, achieving an F1-score of 78.95\%. This improvement over the baseline indicates that self-attention can effectively model features from different extractors and focus on relevant parts of the input sequence.
Finally, the proposed model, which incorporates both cross-attention and self-attention, achieved the best overall performance, with an accuracy of 75.19\% and an F1-score of 79.13\%. This superior performance supports our hypothesis that pitch information, when fused with deep learning features through cross-attention, provides valuable additional context for the task. The combination of cross-attention for inter-feature fusion and self-attention for intra-feature attention enables the model to effectively leverage both types of relationships, resulting in the highest recall and a balanced performance across all metrics.


\subsection{Comparative experiments}
\subsubsection{Private dataset}
Table~\ref{tab:result:private} reports our performance of the comparison models on the private dataset from psychological support hotline emotion analysis task. 
The proposed model achieves the best performance in F1-score. While Wav2Vec 2.0 demonstrates strong performance compared to other models, particularly achieving the highest accuracy among the comparison models, it performs lower than several models in terms of F1-score. This is significant because F1-score, as a balanced metric that considers both precision and recall, provides a more reliable measure for evaluating performance on imbalanced datasets.
The lower performance in F1-score may be due to its reliance on raw audio representations without leveraging additional features like pitch or MFCCs, which are crucial for emotion recognition. Whisper-small-Chinese-base, optimized for Chinese language tasks, performed well with an accuracy of 76.23\%, while other Whisper models (medium and large-v3) showed comparable results but still fell short of Wav2Vec 2.0. HuBERT had the lowest accuracy among the pre-trained models.
Thanh et al. (2024) proposed a method combining Wav2Vec 2.0 features with pitch features through cross-attention and self-attention mechanisms. Their approach showed an improvement in recall (+1.16\% point) and F1-score (+0.85\% point), highlighting the effectiveness of integrating pitch features and attention mechanisms for enhanced emotion recognition in speech. While their accuracy was similar to Wav2Vec 2.0, the improvement in recall demonstrated the advantage of combining multiple feature types for emotion detection.
Our proposed method, integrating Wav2Vec 2.0 with pitch and MFCC features using attention mechanisms, achieved an F1 score of 79.13\%. It outperformed all models in recall and F1-score, demonstrating superior emotional sensitivity. The integration of diverse feature types and attention mechanisms allowed our method to better capture complex emotional cues in speech, leading to an improved recall and a balanced F1-score.

\begin{table}[ht]
    \centering
    \caption{Performance results on the private dataset from the Chinese psychological support hotline.}
    \begin{tabular}{llll} 
    \hline
    Models                     & Accuracy & Recall  & F1-score  \\ 
    \hline
    Wav2Vec 2.0                & \textbf{76.97\%}  & 76.96\% & 76.96\%   \\
    HuBERT                     & 73.29\%  & 73.23\% & 73.23\%   \\
    Whisper-small              & 74.94\%  & 74.92\% & 74.92\%   \\
    Whisper-small-Chinese-base & 76.23\%  & 76.03\% & 76.03\%   \\
    Whisper-medium             & 75.81\%  & 75.58\% & 75.58\%   \\
    Whisper-large-v3           & 75.56\%  & 75.52\% & 75.52\%   \\
    Thanh et al.\cite{thanh2024robust}                & 76.20\%  & 78.12\% & 77.81\%   \\
    \textbf{The proposed}               & 75.19\%  & \textbf{87.75\%} & \textbf{79.13\%}   \\
    \hline
    \end{tabular}
    \label{tab:result:private}
\end{table}

\subsubsection{Public dataset}
We selected the comparably performing model, Wav2Vec 2.0, for comparison on the public ViSEC dataset, with the results reported in Table~\ref{tab:result:public}. 
The proposed model achieves the best performance across both evaluation metrics, further validating the findings from the experiments on the private dataset, as outlined in Table~\ref{tab:result:private}. The use of a single feature from Wav2Vec proves insufficient to capture the complexity of emotions. In contrast, the results from Thanh et al. \cite{thanh2024robust} demonstrate approximately a 10\% points of improvement in weighted accuracy over the Wav2Vec model, attributed to the integration of deep learning features with tonal features. Ultimately, our proposed model achieves superior performance due to the incorporation of attention mechanisms and the combination of MFCC features. This highlights the robustness of our model in capturing negative emotions, making it well-suited for emotion recognition tasks.

\begin{table}[ht]
    \centering
    \caption{Performance results on the public Visec dataset. ``UA'' represents unweighted accuracy, and ``WA'' represents the weighted accuracy.}
    \begin{tabular}{@{} l c c  @{}} 
        \toprule 
        Method & UA & WA  \\ 
        \midrule 
        Wav2Vec 2.0~\cite{thanh2024robust}      & 61.74\%  &   62.37\%  \\
        Thanh et al.~\cite{thanh2024robust}     & 72.72\%  &   72.72\%    \\
        \textbf{The proposed}     & \textbf{73.30\%}   &   \textbf{73.30\%}   \\
        
        \bottomrule 
    \end{tabular}
    \label{tab:result:public}
\end{table}
Overall analysis of the results from both experiments shows that our model, through the feature fusion strategy, excels in both recall and classification performance, fully demonstrating its application potential in emotion recognition tasks. This not only emphasizes the importance of diverse feature selection but also provides theoretical support for subsequent research.

\subsection{Emotion trend analysis}
\begin{figure*}[!ht] \centering\includegraphics[width=1\linewidth]{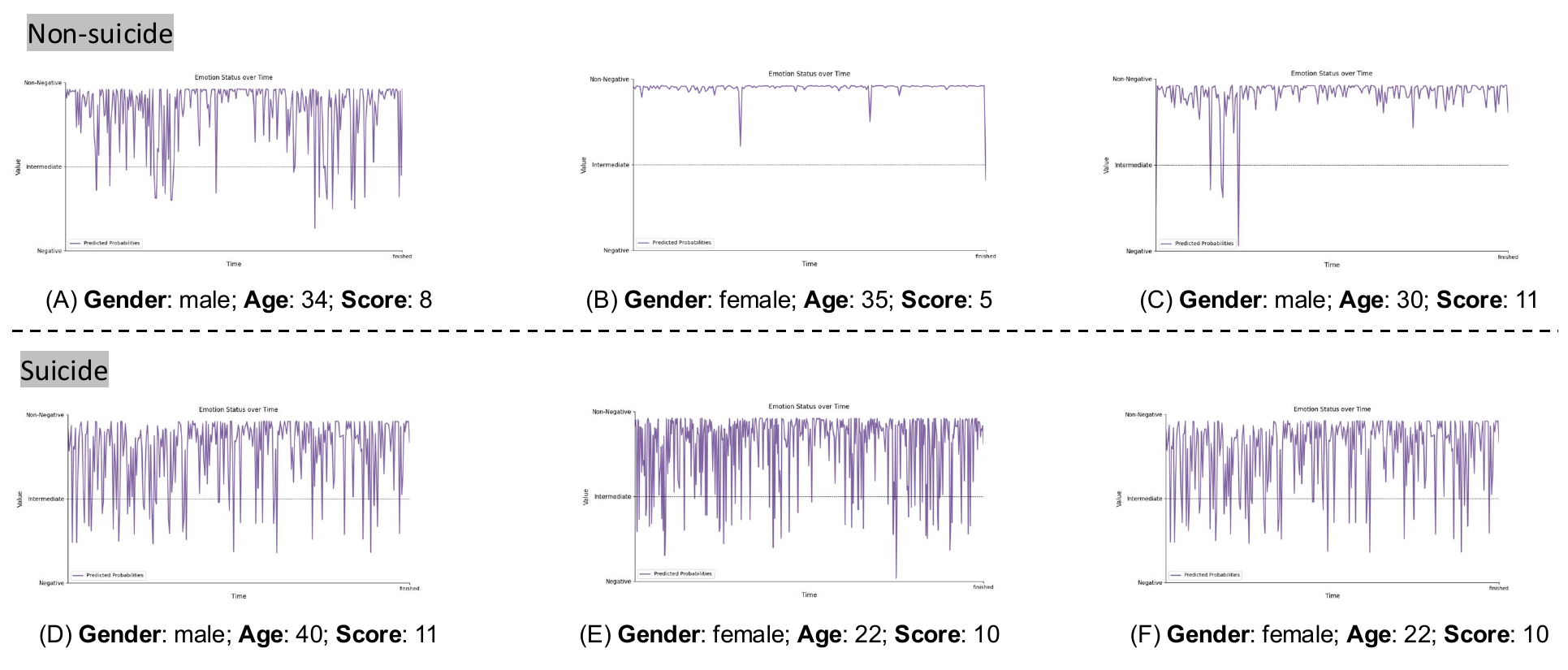}
    \caption{Emotional trends (from Negative to Non-negative) of callers during the full conversation. The ``Non-suicide'' group (A–C) includes individuals without suicidal behavior during the 12-month follow-up, while the ``Suicide'' group (D–F) includes those with suicide behavior. 
    The ``Score'' represents the mental health scale ratings assessed during the hotline calls. }
    \label{fig:emotion_trend} 
\end{figure*}

We used the trained model to analyze emotional trend changes across suicide and non-suicide behavior groups. Six examples (three suicide subjects and three non-suicide subjects) are presented in Figure~\ref{fig:emotion_trend}.
The result can be seen in Table~\ref{tab:result_emotion}.

\begin{table}[ht]
\centering
\caption{The mean values and 95\% bootstrap confidence intervals of negative speech segments (NSS) and emotional change rate (ECR) across the full counseling stage and the suicide assessment stage (first 30 minutes).}
\label{tab:result_emotion}
\begin{tabular}{|c|c|c|c|} 
\hline
Group                        & Stage      & NNS (95\% BCI) & ECR (95\% BCI)      \\ 
\hline
\multirow{2}{*}{Suicide}     & Assessment & 24 [17, 31]   & 0.15 [0.12, 0.22]  \\ 
\cline{2-4}
                             & Full       & 37 [26, 49]   & 0.17 [0.12, 0.22]  \\ 
\hline
\multirow{2}{*}{Non-suicide} & Assessment & 21 [12, 30]   & 0.13 [0.09, 0.20]  \\ 
\cline{2-4}
                             & Full       & 30 [20, 42]   & 0.13 [0.09, 0.16]  \\
\hline
\end{tabular}
\end{table}


The results revealed that, on average, the suicidal group exhibited higher values for both NSS and ECR in comparison to the Non-suicide group across both stages. During the suicide assessment stage, the mean NSS in the suicidal group was 24 [17, 31], slightly higher than the Non-suicide group’s mean of 21 [12, 30]. In the full counselling stage, the NSS mean values were 37 [26, 49] for the suicidal group and 30 [20, 42] for the Non-suicide group. However, the differences in NSS between the two groups did not reach statistical significance in either stage as indicated by p-values of 0.77 and 0.38 for the suicide assessment and full counselling stages, respectively.
A similar trend was observed for ECR. In the suicide assessment stage, the suicidal group had a mean ECR of 0.15 [0.12, 0.22], slightly higher than the Non-suicide group’s 0.13 [0.09, 0.20]. During the full counselling stage, the mean ECR values were 0.17 [0.12, 0.22] for the suicidal group and 0.13 [0.09, 0.16] for the Non-suicide group. Despite these apparent differences, the p-values of 0.59 for the assessment stage and 0.22 for the full counselling stage suggest that these differences are not statistically significant.
These findings are further illustrated in Figure~\ref{fig:emotion_trend}, where the suicidal group (A, B, and C) exhibits more frequent emotional changes compared to the non-suicidal group (D, E, and F). The non-suicidal group exhibited fewer and less intense emotional fluctuations overall, with Subject (B) showing the most stable trajectory, while Subject (C), despite a high score of 11, displayed stable emotional trends. In contrast, the suicidal group showed more frequent and pronounced fluctuations, with Subjects (D), (E), and (F) exhibiting heightened emotional instability. Notably, a general relationship was observed between higher assessment scores and greater emotional variability, though Subject (C) emerged as an exception. These findings suggest that emotional fluctuation patterns could serve as a valuable metric to differentiate between suicidal and non-suicidal individuals. However, the case of Subject (C) highlights the need to consider additional factors, such as age, gender, and other contextual variables, to refine the relationship between emotional trends and assessment scores in future research.

Overall, while the suicidal group displayed higher mean values for NSS and ECR across both stages, the overlapping confidence intervals and non-significant p-values indicate that these metrics do not effectively differentiate between the Suicide and Non-suicide groups in the present dataset. Future studies may consider exploring additional metrics, larger sample sizes, or more granular analyses to identify factors that could better distinguish these groups.

\section{Discussion}
In this paper, we developed and evaluated a novel speech emotion recognition model. Utilizing data from China's largest mental health counseling hotline, our model achieved an F1-score of 79\% on the negative emotion analysis task, outperforming other state-of-the-art (SOTA) models. Additionally, we trained and evaluated our model on public datasets, where it also demonstrated superior performance. Using this model, we conducted an emotion analysis on external data to compare emotional trend changes between suicidal and non-suicidal groups. The results revealed that the suicidal group exhibited a higher frequency of negative emotions and more rapid emotional fluctuations. These findings underscore the heightened emotional instability in the suicidal group and demonstrate the potential of our approach for clinical applications.

Compared with related deep learning models for semantic analysis in audio data~\cite{radford2023robust}, ~\cite{thanh2024robust}, ~\cite{baevski2020wav2vec}, ~\cite{hsu2021hubert}, our model integrates information from diverse features, making it better suited to the unique characteristics of audio analysis tasks in the psychological support hotline context. Our primary focus is on applying the model in clinical settings, which underscores its practical value for real-world applications.
The findings of this study highlight the potential of speech emotion recognition models in identifying emotional instability, particularly in populations at risk of suicide. 

However, our research has certain limitations. Although the results demonstrated that the suicidal group displayed more frequent negative emotions and greater emotional fluctuations than the non-suicidal group, the lack of statistical significance during the assessment stage suggests the need for further refinement in both the analytical methods and the modeling approach. Several factors could be explored in future research to address these limitations and expand the applicability of our model.  
First, gender differences in emotional expression should be considered, as men and women often exhibit distinct patterns of emotional regulation. Incorporating gender as a variable could provide a more nuanced understanding of emotional trends in suicidal individuals and improve the model's precision. Additionally, the binary classification of suicidal and non-suicidal groups may oversimplify the complexity of suicidal behavior. Future studies could refine this categorization by considering short-term (e.g., 24-hour) and medium-term (e.g., 3-month) risks, which might reveal temporal dynamics crucial for early intervention.  
Moreover, the integration of emotion trend metrics with psychometric scale scores could further validate our findings and enhance clinical relevance. For instance, correlating emotional fluctuations with depression or anxiety scores may uncover deeper associations between emotional variability and mental health states. Finally, the emotional change rate itself holds promise as a novel metric for mental health assessment. As emotional variability is particularly pronounced in adolescents compared to adults, age-stratified analyses could explore whether emotional fluctuations are influenced by developmental stages, thus providing targeted insights for different age groups.  
These considerations underscore the need for a broader and more detailed exploration of emotional trends, paving the way for the development of refined tools to support mental health care and suicide prevention efforts.

\section{Conclusion}
In summary, we developed a speech emotion recognition model using data from China's largest psychological support hotline. The proposed model demonstrated high accuracy, achieving the best performance compared to eight other models. It was also validated on a public dataset, where it outperformed state-of-the-art models. Applying the model in a clinical setting, we analyzed emotional trends between suicidal and non-suicidal individuals. Our findings revealed that the suicidal group exhibited a higher emotional change rate and more frequent emotional fluctuations. The proposed model shows promise as a valuable analytical tool for clinical research and has the potential to aid in timely interventions for individuals with suicidal tendencies.

\section{Acknowledgments}
This study was supported by the National Natural Science Foundation of China, Beijing Municipal High Rank Public Health Researcher Training Program [2022-2-027], the Beijing Hospitals Authority Clinical Medicine Development of Special Funding Support [ZYLX202130], and the Beijing Hospitals Authority’s Ascent Plan [DFL20221701]. 
Guanghui Fu is supported by a Chinese Government Scholarship provided by the China Scholarship Council (CSC).

\bibliographystyle{IEEEtran}
\bibliography{ref}
\end{document}